\PassOptionsToPackage{unicode}{hyperref}
\PassOptionsToPackage{hyphens}{url}
\PassOptionsToPackage{dvipsnames,svgnames,x11names}{xcolor}
\documentclass[
]{article}
\usepackage{xcolor}
\usepackage{amsmath,amssymb}
\usepackage{iftex}
\ifPDFTeX
  \usepackage[T1]{fontenc}
  \usepackage[utf8]{inputenc}
  \usepackage{textcomp} 
\else 
  \usepackage{unicode-math} 
  \defaultfontfeatures{Scale=MatchLowercase}
  \defaultfontfeatures[\rmfamily]{Ligatures=TeX,Scale=1}
\fi
\usepackage{lmodern}
\ifPDFTeX\else
\fi
\IfFileExists{upquote.sty}{\usepackage{upquote}}{}
\IfFileExists{microtype.sty}{
  \usepackage[]{microtype}
  \UseMicrotypeSet[protrusion]{basicmath} 
}{}
\makeatletter
\@ifundefined{KOMAClassName}{
  \IfFileExists{parskip.sty}{%
    \usepackage{parskip}
  }{
    \setlength{\parindent}{0pt}
    \setlength{\parskip}{6pt plus 2pt minus 1pt}}
}{
  \KOMAoptions{parskip=half}}
\makeatother
\makeatletter
\ifx\paragraph\undefined\else
  \let\oldparagraph\paragraph
  \renewcommand{\paragraph}{
    \@ifstar
      \xxxParagraphStar
      \xxxParagraphNoStar
  }
  \newcommand{\xxxParagraphStar}[1]{\oldparagraph*{#1}\mbox{}}
  \newcommand{\xxxParagraphNoStar}[1]{\oldparagraph{#1}\mbox{}}
\fi
\ifx\subparagraph\undefined\else
  \let\oldsubparagraph\subparagraph
  \renewcommand{\subparagraph}{
    \@ifstar
      \xxxSubParagraphStar
      \xxxSubParagraphNoStar
  }
  \newcommand{\xxxSubParagraphStar}[1]{\oldsubparagraph*{#1}\mbox{}}
  \newcommand{\xxxSubParagraphNoStar}[1]{\oldsubparagraph{#1}\mbox{}}
\fi
\makeatother

\usepackage{longtable,booktabs,array}
\usepackage{calc} 
\usepackage{etoolbox}
\makeatletter
\patchcmd\longtable{\par}{\if@noskipsec\mbox{}\fi\par}{}{}
\makeatother
\IfFileExists{footnotehyper.sty}{\usepackage{footnotehyper}}{\usepackage{footnote}}
\makesavenoteenv{longtable}
\usepackage{graphicx}
\makeatletter
\newsavebox\pandoc@box
\newcommand*\pandocbounded[1]{
  \sbox\pandoc@box{#1}%
  \Gscale@div\@tempa{\textheight}{\dimexpr\ht\pandoc@box+\dp\pandoc@box\relax}%
  \Gscale@div\@tempb{\linewidth}{\wd\pandoc@box}%
  \ifdim\@tempb\p@<\@tempa\p@\let\@tempa\@tempb\fi
  \ifdim\@tempa\p@<\p@\scalebox{\@tempa}{\usebox\pandoc@box}%
  \else\usebox{\pandoc@box}%
  \fi%
}
\def\fps@figure{htbp}
\makeatother

\NewDocumentCommand\citeproctext{}{}

\makeatletter
 \let\@cite@ofmt\@firstofone
 \def\@biblabel#1{}
 \def\@cite#1#2{{#1\if@tempswa , #2\fi}}
\makeatother
\newlength{\cslhangindent}
\newlength{\csllabelwidth}
\newenvironment{CSLReferences}[2] 
 {\begin{list}{}{%
  \setlength{\itemindent}{0pt}
  \setlength{\leftmargin}{0pt}
  \setlength{\parsep}{0pt}
  \ifodd #1
   \setlength{\leftmargin}{\cslhangindent}
   \setlength{\itemindent}{-1\cslhangindent}
  \fi
  \setlength{\itemsep}{#2\baselineskip}}}
 {\end{list}}
\usepackage{calc}

\usepackage{caption}
\usepackage{float}
\usepackage{newunicodechar}
\newunicodechar{α}{\ensuremath{\alpha}}
\newunicodechar{Δ}{\ensuremath{\Delta}}
\usepackage{caption}
\makeatletter
\@ifpackageloaded{caption}{}{\usepackage{caption}}
\AtBeginDocument{%
\ifdefined\contentsname
  \renewcommand*\contentsname{Table of contents}
\else
  \newcommand\contentsname{Table of contents}
\fi
\ifdefined\listfigurename
  \renewcommand*\listfigurename{List of Figures}
\else
  \newcommand\listfigurename{List of Figures}
\fi
\ifdefined\listtablename
  \renewcommand*\listtablename{List of Tables}
\else
  \newcommand\listtablename{List of Tables}
\fi
\ifdefined\figurename
  \renewcommand*\figurename{Fig.}
\else
  \newcommand\figurename{Fig.}
\fi
\ifdefined\tablename
  \renewcommand*\tablename{Table}
\else
  \newcommand\tablename{Table}
\fi
}
\@ifpackageloaded{float}{}{\usepackage{float}}
\floatstyle{ruled}
\@ifundefined{c@chapter}{\newfloat{codelisting}{h}{lop}}{\newfloat{codelisting}{h}{lop}[chapter]}
\floatname{codelisting}{Listing}

\makeatother
\makeatletter
\@ifpackageloaded{caption}{}{\usepackage{caption}}
\@ifpackageloaded{subcaption}{}{\usepackage{subcaption}}
\makeatother
\usepackage{bookmark}
\IfFileExists{xurl.sty}{\usepackage{xurl}}{} 
\hypersetup{
  pdftitle={Who is the Pronoun Pro?},
  pdfauthor={Carlos Hartmann},
  colorlinks=true,
  linkcolor={blue},
  filecolor={Maroon},
  citecolor={Blue},
  urlcolor={Blue},
  pdfcreator={LaTeX via pandoc}}

\title{Who is the Pronoun Pro?}
\usepackage{etoolbox}
\makeatletter
\providecommand{\subtitle}[1]{
  \apptocmd{\@title}{\par {\large #1 \par}}{}{}
}
\makeatother
\subtitle{Comparing LLMs and Humans at Number-Disambiguation of
\emph{they}}
\author{Carlos Hartmann\\{\normalsize University of
Zurich}\\{\normalsize\href{mailto:carlos.hartmann@es.uzh.ch}{carlos.hartmann@es.uzh.ch}}\\{\normalsize ORCID: \href{https://orcid.org/0009-0007-6930-2284}{0009-0007-6930-2284}}}
\date{}
\begin{document}
\maketitle

\newpage

\section*{Abstract}\label{abstract}
\addcontentsline{toc}{section}{Abstract}

This study evaluates whether large language models (LLMs) can reliably
replace or complement human research assistants in linguistically
complex annotation. In a controlled annotation task, 73 human annotators
(local students and crowdworkers with and without a background in
linguistics) and more than 40 LLM configurations number-disambiguated
174 authentic uses of \emph{they} from Reddit, including innovative and
context-sensitive cases. The best LLM, \emph{Kimi K3}, reached 94.5\%
accuracy, above the human majority vote (86.2\%) and the mean of the top
half of human annotators (78.5\%); only the single best human annotator
scored higher (95.7\%). LLMs also offer substantial practical advantages
in cost and speed. However, qualitative error analysis reveals a crucial
complementarity: human reasoning in low-agreement cases exposes
inconsistencies in the annotation typology and thereby improves the
methodological framework itself. The study contributes (i) a
benchmark-style human--LLM comparison for a semantically demanding task,
(ii) transparent cost and time evidence, and (iii) a methodological
argument against a purely replacement-oriented view of automation. LLMs
are viable for large-scale annotation support in linguistics, but their
responsible use requires human critical oversight. The dataset with its
gold-standard labels, all human and LLM annotations, and the fully
reproducible code are openly available as a Quarto-based, public
repository.

Keywords: large language models; reasoning models; corpus annotation;
singular \emph{they}; AI ethics

\newpage

\section{Introduction}\label{sec-introduction}

Large Language Models (LLMs) are changing the methodological landscape
in the humanities and social sciences, and linguistics more
specifically. The promise of the team developing \emph{GPT-3} was that
Natural Language Processing (NLP) as a whole would shift away from
models specially-created for specific tasks ``to using task-agnostic
pre-training and task-agnostic architectures'' (Brown et al., 2020, p.
1), i.e.~general-purpose language models able to perform a wide range of
tasks without task-specific training. Almost instantly, LLMs began to
find their way into different areas of linguistics research, among them
morphological pattern matching (Jubelius, 2025), authorship verification
(Huang \& Grieve, 2024), semantics-based corpus extraction (Divjak et
al., 2024), and the annotation of pragmatic and discourse features (Yu
et al., 2024).

The present work is the continuation of an older, anticipatory effort
(Hartmann, 2022), which tested the then-new \emph{GPT-3} for its ability
to disambiguate plural and singular uses of the pronoun \emph{they}.
This is, in essence, a coreference resolution (CRR) and classification
task, as the number-disambiguation of \emph{they} hinges on the
pronoun's referent\footnote{The linguistic element to which a pronoun
  points is normally called the `antecedent'. The `referent' denotes the
  real-world entity behind that antecedent. Since pronouns can sometimes
  have a referent without an antecedent appearing in the utterance, and
  it is the referent that truly number-disambiguates the meaning of
  \emph{they}, the term `referent' is preferred in this paper.}. CRR as
a field precedes LLMs (see Gan et al., 2025; Jurafsky \& Martin, 2026),
though it struggled with the persistent challenges posed by the nature
of coreference, such as real-world knowledge required for textual
understanding and context-dependency (Novák et al., 2025; Seminck et
al., 2025). What is more, English pronoun use has shifted in recent
years (Newman, 1997; Paterson, 2014), most notably with the extension of
\emph{they} as a personal pronoun for people of non-binary gender
(Conrod, 2019), but also with the growing visibility of neologistic
pronouns (Baron, 2020; Remsö et al., 2026; Rose et al., 2023). Such
innovative language posed yet another challenge that established CRR
techniques struggled to overcome (Cao \& Daumé III, 2020). LLMs, on the
other hand, showed promise in their context-sensitivity and advanced
textual understanding (Brown et al., 2020; Pavlovic \& Poesio, 2024),
and would perhaps also overcome the challenge of language innovations
such as the one posed by \emph{they}. This hope was at first not
confirmed in an early study (Hartmann, 2022, Section 5.6), but testing
continued to monitor progress in LLM performance. When Anthropic's LLM
\emph{Claude Sonnet} reached both recall and precision of over 80\% on
the same dataset in late 2024, it seemed plausible that LLMs may have
reached human-like performance in this task. In order to assess this
performance reliably and accurately, the present study focuses on this
research question: ``To what extent are number-annotations of authentic
uses of \emph{they} by LLMs reliable when compared to trained human
annotators?'' Beyond the case of \emph{they}, the answer bears on how
annotated language resources are built: whether, and for which cases,
LLMs can take over annotation work from trained human annotators.

This research is situated in the broader context of LLMs expanding in
uses far beyond the original purpose of solving one-off tasks. Through
the development of techniques like hidden prompts, chain-of-thought
prompting, and reasoning models more generally, LLMs have attained the
status of assistants who take over entire workflows. They perform
research assistance work at the level of a typical research intern, if
not in output quality, then certainly in monetary cost and speed --
although the true cost of LLMs goes beyond the cost of tokens. Still,
the consequences of this technology on workflows in academia, and
elsewhere, especially as concerns the role of human labor, are likely
significant, though yet unclear. The present research is thus not only
about the performance of LLMs on a specific task, but about the broader
question on the viability of employing them as one would human research
assistants. The study is conceived as a simulation of handing off an
annotation task to a student assistant, reflected in the recruitment
strategy and task design. As the assessment of annotation quality
demands more nuance than a simple scored performance can provide, this
study investigates both quantitative and qualitative measures.

Three falsifiable, quantitative hypotheses were formulated at the
outset, addressing three levels of performance:

\begin{quote}
Hypothesis 1: the most accurate number-disambiguator of \emph{they} will
be an LLM.

Hypothesis 2: the most accurate LLM at number-disambiguating \emph{they}
will perform at a significantly higher accuracy than the mean human
accuracy.

Hypothesis 3: the performance of the most accurate LLM will not be
significantly worse at rare and innovative uses of \emph{they} than the
mean human performance.
\end{quote}

The hypotheses form a cascade: should Hypothesis 1 not be confirmed,
Hypothesis 2 would still attest that LLMs largely outperform humans;
should that fail as well, Hypothesis 3 would still justify the use of
LLMs for semantically complex tasks such as the resolution of innovative
pronouns.

``Rare and innovative uses of \emph{they}'' in Hypothesis 3 are defined
as uses in which the referent of \emph{they} is one specific person both
semantically and grammatically, i.e.~`specific singular \emph{they}'
(see Section~\ref{sec-kinds-of-they}).

\section{\texorpdfstring{Coreference resolution and the changing pronoun
\emph{they}}{Coreference resolution and the changing pronoun they}}\label{coreference-resolution-and-the-changing-pronoun-they}

This study draws on research from two main areas: NLP research on CRR
and linguistics research on the history of change of the pronoun
\emph{they}. Both inform the design of the task, and results will be
relevant to both topics. More broadly, however, this research is
relevant for any task in linguistics that surpasses the written surface
level and is thus dependent on the availability of trained human
annotators.

\subsection{CRR as performed by LLMs}\label{sec-coreference-resolution}

The task of CRR requires a deep understanding of the text. The first
step is to identify all `mentions' of real-world entities (i.e.~specific
linguistic material denoting or even implying entities) and then to
identify which of these mentions refer to the same entity, i.e.~which of
these mentions `corefer' (Jurafsky \& Martin, 2026, p. 1). As with many
tasks in NLP, early rule-based approaches gave way to machine learning
(ML)-based approaches (Gan et al., 2025). The most complete history of
automatic CRR is provided in Seminck et al. (2025). Current research
increasingly involves LLMs, though these have yet to outperform
traditional approaches across the board (Novák et al., 2025, Section 5).

Traditional approaches are computationally cheaper than LLMs (Seminck et
al., 2025, Section 3.1.2), lending them a significant advantage in many
applications. Their disadvantages, however, have proven difficult to
overcome: ML systems require highly specialized training material, which
is not only scarce but often varies in annotation standards, making
generalizable training difficult (Seminck et al., 2025, Section 3.1.2).
This makes it difficult to expand to other languages or to language
innovations such as singular \emph{they}, which are absent from popular
datasets such as OntoNotes (Webster et al., 2018, Section 2.1;
Weischedel et al., 2013). Like in other areas of NLP, LLMs\footnote{The
  term `Large Language Model' denotes a variety of different model
  architectures. In the present work, the term denotes generative,
  decoder-only models.} offer a great deal of flexibility at a
remarkably high cost of computation.

State-of-the-art research on LLM-based CRR was presented at the CRAC
2025 workshop of the Association for Computational Linguistics, whose
shared task on multilingual CRR had separate tracks for traditional and
LLM-based systems. Traditional approaches clearly won (Novák et al.,
2025, Section 5), and Seminck et al. (2025, Section 4) contributed the
winning LLM system.

Seminck et al. (2025) frame context management as a central bottleneck:
shorter prompts worked better, plausibly because they leave more room
for input context and, in my interpretation, reduce over-determination
of the task, while overly long inputs led to ``disastrously bad''
outcomes, with the model appearing to lose track of the objective (2025,
Section 3.2.2). Adding information is thus often necessary, but careful
context window management remains critical (2025, Section 5). The
trade-off is also practical: a traditional pipeline processed their data
in 16 minutes, the LLM-based approach in roughly 1.5 days (2025, Section
4). Novák et al. (2025) nonetheless see clear potential for LLMs on
tasks that require a broad spectrum of real-world knowledge.

Whether to prompt or to fine-tune LLMs remains an open methodological
question: the winning LLM system was fine-tuned (Seminck et al., 2025),
but the second-best was purely prompting-based and only 1\% lower in F1
(Novák et al., 2025, Section 5). The present study relies on prompting,
as all-purpose, out-of-the-box LLMs are more directly comparable to
human annotators, who also encounter task instructions and data without
prior task-specific adaptation.

To my knowledge, only one other study compared the performance of LLMs
and humans on a CRR task. Using the AmbiCoref dataset of
template-generated minimal pairs, Shore et al. (2025) demonstrate a
persistent trade-off: while LLMs can achieve high accuracy on
unambiguous sentences or strong ambiguity detection, they cannot achieve
both simultaneously. Crucially, they found that models fail to shift
their response patterns between ambiguous and unambiguous text samples
as humans naturally do. This means that models fare well at recognizing
ambiguity when this is what is asked of them; but when the task is to
find coreference, they will force an answer without considering first if
the language might be ambiguous and thus not have a clear coreference in
the first place. The present study will likewise demand different
behavior depending on ambiguity of the input.

Outside CRR, Yu et al. (2024) offered one of the first direct
comparisons of an LLM and a human coder in corpus linguistics,
annotating apologies -- like the number of \emph{they}, a
discourse-level phenomenon that hinges on language understanding in
context rather than on form. \emph{GPT-4} came close to the human coder
(92.7\% vs.~95.4\% accuracy), though both struggled with categories
realized by a diverse range of forms, and the human clearly outperformed
the model in recognizing uses of \emph{sorry} that were not apologies
(2024, Section 4). Morin \& Marttinen Larsson (2026) have since scaled
up such annotation to over 140,000 concordance lines from COHA with
\emph{GPT-5}, at 98--99\% accuracy against their own manual annotation.

\subsection{\texorpdfstring{Three kinds of
\emph{they}}{Three kinds of they}}\label{sec-kinds-of-they}

The typology of \emph{they} used in this study is that of Hartmann
(2026a), differentiating between plural, generic singular, and specific
singular \emph{they}. Plural \emph{they} refers to a group of people.
Generic singular \emph{they} refers to multiple potential referents via
a singular expression and thus lacks a specific referent (e.g.~\emph{any
expert worth their salt}); in the research literature, it is often
simply called `singular \emph{they}'. Specific singular \emph{they},
also known as `referential \emph{they}', refers to a specific individual
whose gender is unknown or non-binary (e.g.~\emph{our lawyer and their
trainees} or \emph{Judith Butler defended their stance}). For the
purposes of this study, I do not differentiate between the
gender-obfuscating use when referring to the hearer-unknown
\emph{lawyer} and the overtly non-binary-gendered use when referring to
Judith Butler. In the present article, the terms `generic \emph{they}'
and `singular \emph{they}' signify generic singular and specific
singular \emph{they}, respectively. Further distinctions could be made
between different kinds of generic and referential usage, see Newman
(1997), but these pertain more to a detailed linguistic analysis than to
a typical annotation task.

\section{Data and methods}\label{data-and-methods}

The present research focuses on a specific aspect of CRR: the
number-disambiguation of \emph{they}. Mention detection in classical CRR
is precluded, as the mention is defined as any instances of the pronoun
\emph{they}. The goal is thus not necessarily to point to the antecedent
in the text, but rather to identify the number of the referent.

\subsection{Task design}\label{sec-task-design}

Participants, human and LLM-based, had to determine the number of the
referent of the pronoun \emph{they} in a given Reddit comment, using the
provided context when necessary, and select the appropriate category
(plural, generic, or singular).

Each participant (see Section~\ref{sec-participants} below) was given a
250-word explanation of my typology of \emph{they} (see
Section~\ref{sec-kinds-of-they})\footnote{The same text is part of the
  LLM prompts, e.g.~\texttt{0-assets/prompting/context\_agnostic.txt} in
  the study repository (Hartmann, 2026b).}, followed by ten example
items in random order, curated so that each category was represented at
least once and a few required looking up the context to disambiguate.
`Context' here refers to the Reddit thread in which a comment was
originally written, not to the text immediately surrounding the pronoun.
The examples used the same interface as the actual annotation, including
access to the Reddit context via URL.

They then assigned each of the 174 Reddit comments (48 for each human
participant) to one of the three categories, with the option to explore
the context whenever they could not immediately identify the referent of
the pronoun. For each item, they could comment on their reasoning
(required for the first ten items, optional thereafter) and indicate if
they felt uncertain of their choice.

Example \hyperref[hiphop-example]{(1)}, an item from the initial sample
sentences, illustrates the task design:

\phantomsection\label{hiphop-example}
\begin{quote}
(1) Not hip-hop related, and probably shouldn't say in case anything
does happen, just kinda bummed and not sure what I did to get
essentially ghosted. Pretty discouraging seeing \emph{them} post on
social media but not reply to their personal email (emphasis mine; item
ID ma\_era\_306411 from the study dataset)
\end{quote}

The text alone only reveals that a person or persons with a social media
account and personal email are no longer responding to the commenter.
The context on Reddit\footnote{Archived version accessible at:
  \url{https://archive.is/SDsNe}.} contains the commenter's prior
comment about ``an artist,'' confirming that the \emph{them} in question
is specific singular.

The task design was identical for LLMs, except that they were not shown
the ten examples, which mostly served to explain the annotation
interface to human participants and would have overloaded the context
window with redundant text. LLMs received the same text explaining the
goal of the task, the availability of the context, and the three
categories of \emph{they} with their definitions and illustrative
examples. It covered the same theory as the ten examples, and LLMs
generally fare well with just one explanation. In lieu of the comment
function, LLMs were prompted not to withhold reasoning, and their full
responses were recorded, though these should not be over-interpreted as
true ``reasoning''. As uncertainty is not truly available to current
LLMs, a drawback of LLM-based CRR discussed by Seminck et al. (2025,
Section 5.4), the present study does not compare uncertainty between
humans and LLMs.

The role of context differed for human annotators and LLMs. For humans,
context-dependent items were assumed to add time without posing a real
challenge, so only a few were included, serving to filter out
inattentive participants (see
Section~\ref{sec-filtered-human-performance}). For LLMs, however,
context-dependency adds relevant complexity, given the apparent
trade-off between accuracy and the recognition of ambiguous cases (Shore
et al., 2025). Asked whether Example \hyperref[hiphop-example]{(1)}
requires further context, an LLM may correctly say so; asked directly
for the referent, it may guess a band or an agency instead of a single
artist rather than first critically assessing knowability. These
trade-off effects are tested in Section~\ref{sec-context-dependence}.

\subsection{Dataset}\label{sec-dataset}

This study uses data from previously unpublished research that randomly
sampled instances of any form of \emph{they}\footnote{\emph{they},
  \emph{them}, \emph{their}, \emph{theirs}, \emph{themself}, and
  \emph{themselves}.} from the social media website Reddit. 69\% of the
dataset was sampled without any subreddit constraints from the two
months of October 2019 and June 2021. Each instance of a form of
\emph{they} was treated as a separate item and considered by the
sampling algorithm. Each item is thus centered around a single instance
of \emph{they} and its surrounding context. 31\% were sampled from
British\footnote{/r/UnitedKingdom and /r/BritishProblems.} and
Australian\footnote{/r/Australia, /r/Sydney, /r/Melbourne, /r/Brisbane,
  and /r/Adelaide.} subreddits from the entire year 2013 and the month
of October 2022. The 2013 and 2022 data was sampled with a length
restriction of 50--450 characters but not the 2019 and 2021 data. The
174 items in this study were sampled to ensure a balance of items
between plural, generic, and singular \emph{they}. Importantly,
perfectly ambiguous instances due to lacking context or unclear pronoun
reference were disregarded for this study. I re-annotated the dataset
according to the modified annotation scheme of this study and computed
the inter-annotator agreement (IAA) between myself and a colleague with
expertise in linguistics. A near-perfect agreement of 0.83 in Cohen's
kappa confirmed the robustness of the annotation scheme.

Reddit data were chosen for the naturalistic and contemporary nature of
their language. Although OntoNotes is often a popular data source, it
does not contain innovative uses of \emph{they} (Webster et al., 2018,
Section 2.1; Weischedel et al., 2013) and is thus unsuitable for this
study. Annotating Reddit comments more accurately simulates authentic
tasks in linguistics research involving jargon and other domain-specific
language and where context may be needed for correct annotation. The
present dataset thus addresses artificiality concerns while remaining
suitable for a task centering on innovative language use.

Social media data are a grey area for research ethics: Reddit comments
are publicly accessible, but their authors may not necessarily have been
aware when signing off on the Terms of Service, and may not truly
consent to the use of their data. The sampling procedures collected no
personal details beyond what Reddit attaches to each comment, and they
targeted no individuals or groups along any identifiable lines; the only
constraint beyond the random sampling of \emph{they} was the broad
geographical one of the British and Australian subreddits. The published
dataset contains no account names, user flairs, timestamps, or links to
the original comments, which were all removed before publication. It
thus meets § 9(2) of the Information and Data Protection Act of the
Canton of Zurich (IDG), which applies to the University of Zurich and
allows personal data to be processed for purposes that do not concern
particular persons if the data are anonymised and the analyses allow no
inferences about the persons concerned. As the annotation task depends
on their exact wording, the comments are quoted verbatim and could in
principle be traced through a web search; the dataset itself, however,
links no comment to its author.

The dataset was divided into four blocks to manage study duration and
mitigate fatigue effects while maintaining large sample size to ensure
statistical power and diversity in the dataset. Each block contained an
overlapping set of examples common to all participants, supplemented
with block-specific items, and example distribution across blocks was
randomized while maintaining a balance of referent types within blocks.
This resulted in four blocks of 48 items, each of which contained at
least 12 examples of plural, 15 examples of generic, and 13 examples of
singular \emph{they}. Four items in each block were context-dependent,
serving as attention checks (see
Section~\ref{sec-filtered-human-performance}). The order of the items
was randomized across participants to avoid order effects\footnote{LLMs
  do not suffer order effects, as they are prompted on a single example
  each time.}. LLMs were shown an additional 28 context-dependent items
(12 plural, 5 generic, 11 singular) to test their ability to handle such
cases (see Section~\ref{sec-context-dependence}); these are excluded in
direct LLM-to-human comparisons in the Results section. The dataset is
available in the study repository as a CSV file (Hartmann, 2026b).

\subsection{Methods}\label{methods}

Both human and LLM annotators were scored for accuracy against the
ground truth, calculated as the proportion of correct annotations over
the total number of annotations. F1 score and Matthews correlation
coefficient (MCC) were additionally considered, but accuracy proved a
suitable comparison metric as the three classes (plural, generic,
singular) are balanced in the dataset. Since natural class distribution
is strongly imbalanced (plural being by far the most common class), MCC
was later calculated on a projected estimation
(Section~\ref{sec-naturalistic-projection}). For the non-inferiority
analysis in Hypothesis 3, the margin was set to Δ = 0.02 (two percentage
points in label-specific accuracy) as a practical tolerance threshold:
smaller differences are treated as operationally negligible for
assistant-like annotation quality in this task, while larger deficits
would be substantively relevant for downstream linguistic analysis.

The present study is designed as a confirmatory and quasi-experimental
study: research questions and hypotheses are pre-specified based on
previous research, with an experimental design that reflects the
practical constraints of working with human participants (i.e.~study
duration and fatigue effects) and LLMs. Each annotation instance was
treated as an item-level trial for scoring, and condition-wise
breakdowns were computed descriptively at the item level. Nearly all
inferential comparisons were made on aggregated participant- or
run-level performance.

Apart from the LLMs under study, I used AI assistants in preparing this
study and article. \emph{GitHub Copilot} and, later, \emph{Claude Code}
(Anthropic) helped check my existing analysis code and write new code
for the analysis and for rendering this article. \emph{Claude Code} also
assisted in editing the manuscript, mainly by condensing passages and
adapting it to journal style requirements. The text was only edited by
LLM-based tools after receiving thorough, paid copy-editing from Jessica
Roady. I reviewed all changes to code and text and take full
responsibility for the content of this article.

\section{Participants}\label{sec-participants}

Two groups of annotators took part in this study, LLMs and humans, which
are described in turn below.

\subsection{LLMs}\label{llms}

In the present study, LLMs are grouped into commercial and open-source
families; within each family, each model-prompting strategy combination
(with three-fold cross-validation) is treated as one participant.

\subsubsection{Prompt design}\label{sec-prompt-design}

The prompts follow the question-answering (QA) template described by Gan
et al. (2024) and formalized in Gan et al. (2025), structured as
{[}Instruction + Sentences + Question{]}: a short instruction, a set of
sentences containing context, and a question targeting a specific
mention. Although prompts with a specially-devised annotation syntax
yield higher coreference scores, the QA template better preserves
language-understanding behavior, as it requires direct textual reasoning
instead of providing mention annotations (Gan et al., 2025, Section
7.2). It is also closest to the vision in Brown et al. (2020) of
general-purpose language models that perform a wide range of tasks with
minimal pre-processing of data.

The QA template in this study included a theoretical explanation with
short examples, the Reddit comment as context, and a question asking for
the type of \emph{they} in the comment\footnote{It is unclear from
  previous literature if this constitutes zero-shot or few-shot
  prompting. The task description is a few-shot prompt in that it
  contains single-word specifications of the typology of \emph{they}
  uses, but it is also a zero-shot prompt in that it does not contain
  any examples of correct classification.}. A simple system message
instructed the models to approach the task as knowledgeable about
linguistics and especially English pronouns, to respond thoughtfully and
precisely, and to remain aware of potentially relevant context without
inferring beyond what was explicitly stated. The model was then prompted
to output one of three categories: ``plural,'' ``generic,'' or
``singular.'' The full prompt is provided in the study repository
(Hartmann, 2026b). As this approach showed great promise in preliminary
tests, it was used for all LLMs without further experimentation on
prompting techniques. This contrasts with earlier work: for Yu et al.
(2024, Section 3.2), prompt engineering was a central step, with
exemplars, their order, and even category labels iteratively adjusted to
steer \emph{GPT-4}. The present results suggest that current reasoning
models depend far less on such careful prompt crafting, although the
question is not settled: even with \emph{GPT-5}, Morin \& Marttinen
Larsson (2026, pp. 10, 16--17) still invested considerable effort in
prompts that anticipate edge cases.

The only variation in prompting strategy concerned access to the Reddit
context, to test the models' ability to handle context-dependent cases
(see Section~\ref{sec-context-dependence}). Models were prompted
context-agnostically (i.e.~without information about the context), with
context ``on demand'' (i.e.~instructed to request it if needed), or with
the context's permalink. For the most recent additions to the study
(\emph{GPT-5.4}, \emph{Claude Sonnet 5.5}, \emph{Claude Haiku 5.5},
\emph{Kimi K3}, and \emph{MiMo V2.6 Pro}), each request also came with a
web search tool to open the permalink, and these models are reported
``with Web Search''. The search was mandatory for \emph{GPT-5.4}, as it
was for the \emph{GPT-5} runs with context access, while the other
models could decide whether to use it. Models were instructed to end
their response with the chosen category as a single word, which allowed
automatic evaluation without structured output libraries.

\subsubsection{Models}\label{models}

The commercial models include the largest available systems from the
\emph{Claude}, \emph{GPT}, and \emph{Gemini} families. For open-source
models, selection was guided by Xu (2026) to prioritize important and
commonly known options; they include \emph{DeepSeek} models
(\emph{DeepSeek Chat}, \emph{DeepSeek V3.1 Terminus}, and \emph{DeepSeek
V3.2 Speciale}), \emph{Qwen} models (\emph{Qwen3}, \emph{Qwen3-VL},
\emph{Qwen3.5}, and \emph{Qwen-Max} variants), \emph{Llama} models
(\emph{Llama 3.1 405B} and \emph{Llama 4 Maverick/Scout}), \emph{Kimi}
models (\emph{Kimi K2}, \emph{K2 Thinking}, \emph{K2.5}, and \emph{K3}),
\emph{GLM} models (\emph{GLM 4.7}, \emph{GLM 4.7 Flash}, and \emph{GLM
5.1}), \emph{GPT-OSS 120B}, \emph{MiMo V2 Flash}, and \emph{MiMo V2.6
Pro}. Open-source models offer reliable replicability as all changes are
clearly recorded in the documentation, while it is never clear exactly
which parameters and hidden steering settings are used when a request is
sent to a commercial provider. Results for all models can be found in
the study repository (Hartmann, 2026b). For concision, only the
best-performing models are reported in the present article.

\subsection{Humans}\label{humans}

Local students with high performance in linguistics classes were
recruited as participants for the pilot study (henceforth `pilots') to
avoid the ethical concerns of crowdsourced labor and to maintain a
controlled environment for evaluating task design and prompting
strategies. All but one pilot participants performed the study under the
same conditions (as concerns task design and duration) as the later
participants and thus their data, except one, are included in the main
analysis. However, nearly all available participants were L2 speakers of
English, an uncommon demographic for experiments in which experts,
typically L1 speakers, are preferred. Therefore, two additional groups
of participants were recruited through Prolific: one of self-identified
linguists who were L1 speakers of English, and another of non-linguist
L1 English speakers. The human annotators are thus either linguists or
non-linguists, and either L1 or L2 speakers of English. Results thus
bear additional methodological questions regarding the comparison of
these demographic groups as well as that of humans and LLMs.

Blocks A--D of the self-identified linguistics experts on Prolific
comprised 11, 12, 10, and 8 participants respectively, with the pilot
group comprising seven participants. The non-linguist control group,
also stratified into the same A--D blocks, comprised six, six, six, and
seven participants respectively. All participants annotated 48 examples,
yielding a study duration of approximately one hour per participant. The
locals were paid a flat rate of 30 CHF each and the Prolific
participants were given an hourly rate of £15, considered very high on
the platform. Participants could take breaks at any point and were able
to visually tell how far along in the study they were\footnote{The used
  tool was \emph{LimeSurvey} and all items were on one page that
  participants scrolled through.}. The order of the three choices
(plural, generic, singular) was fixed. The study pilots conducted the
survey via \emph{Microsoft Excel} in which I had specially built a
template to easily read and annotate the items (included in the study
repository, Hartmann, 2026b).

\subsubsection{Demographics}\label{demographics}

All participants, whether recruited locally or through Prolific, had to
be female and between the ages of 20 and 27 to choose the participants
most likely to excel at this task: it is well established that young
women are the most innovative speakers (Nevalainen, 2000, p. 38;
Tagliamonte, 2012, pp. 62--63), more aware of gender in language, and
more likely to use singular \emph{they} (Laitinen, 2007, p. 236). Such
sampling also controls for the potential confounds of age and gender. An
additional filter for L1 English and education in ``English language''
(see next section) were also set, depending on study group.

\subsubsection{Expertise}\label{expertise}

The greatest conceptual challenge of the study lies in the comparison of
LLMs with hypothetical student assistants, who are usually selected
purposively: they are screened for academic performance and then chosen
through an application and interview process. The present study likewise
uses a purposive, non-probability sample, designed to approximate the
participants a researcher might select for an annotation task rather
than to represent student assistants statistically. Local participants
were chosen based on their availability and willingness to participate,
the Prolific participants by the criteria described above. These choices
introduce selection bias, but they reflect the study's focus on
potentially suitable human annotators rather than on population-level
human performance. On Prolific, expertise was established through
self-identification as having skills in ``English language,'' which was
assumed to mean linguistics, as the only other comparable category read
``English literature.'' The category is broad, however, and the
demographic requirements left too few participants to further restrict
the pool to those with a linguistics degree, though several had
evidently studied it judging by their comments. Expertise was therefore
also established post hoc (see
Section~\ref{sec-filtered-human-performance}).

\section{Results}\label{results}

The results are reported by hypothesis in Sections
\ref{sec-hypothesis-1}--\ref{sec-hypothesis-3}, followed by additional
metrics in Section~\ref{sec-additional-metrics}.

\subsection{Hypothesis 1}\label{sec-hypothesis-1}

The first hypothesis tests if any LLM-based approach would yield the
highest overall performance. As the results in
Table~\ref{tbl-top10-accuracy} demonstrate, the highest-performing
participant was, in fact, not an LLM. Table~\ref{tbl-top10-accuracy}
shows accuracy, which is a measure of the proportion of correct
responses. However, various LLMs fill the next few places,\footnote{The
  longer this article was in preparation and the more LLMs could be
  tested, the more LLMs reached these places and pushed humans out of
  the top ten. Only first place remained unbeaten at the time of
  submission.} outperforming all 72 of the other human annotators.

\begin{longtable}[]{@{}
  >{\raggedleft\arraybackslash}p{(\linewidth - 4\tabcolsep) * \real{0.1000}}
  >{\raggedright\arraybackslash}p{(\linewidth - 4\tabcolsep) * \real{0.5000}}
  >{\raggedright\arraybackslash}p{(\linewidth - 4\tabcolsep) * \real{0.1800}}@{}}
\caption{The ten highest-performing participants by accuracy. Human
participants are named according to their block and participant number,
while LLM participants are named according to their LLM and prompting
strategy.}\label{tbl-top10-accuracy}\tabularnewline
\toprule\noalign{}
\begin{minipage}[b]{\linewidth}\raggedleft
Rank
\end{minipage} & \begin{minipage}[b]{\linewidth}\raggedright
Participant
\end{minipage} & \begin{minipage}[b]{\linewidth}\raggedright
Accuracy
\end{minipage} \\
\midrule\noalign{}
\endfirsthead
\toprule\noalign{}
\begin{minipage}[b]{\linewidth}\raggedleft
Rank
\end{minipage} & \begin{minipage}[b]{\linewidth}\raggedright
Participant
\end{minipage} & \begin{minipage}[b]{\linewidth}\raggedright
Accuracy
\end{minipage} \\
\midrule\noalign{}
\endhead
\bottomrule\noalign{}
\endlastfoot
1 & Human Pilot Nr. 5 & 95.7\% \\
2 & Kimi K3 with Web Search & 94.5\% \\
3 & Gemini 3 Pro Preview with Context Access & 93.6\% \\
4 & GPT-5 Context on Demand & 93.1\% \\
5 & Claude Sonnet 5.5 with Web Search & 91.7\% \\
6 & MiMo V2.6 Pro with Web Search & 91.0\% \\
7 & Claude Haiku 5.5 with Web Search & 90.8\% \\
8 & GPT-5 Context-Agnostic & 90.7\% \\
9 & GPT-5 with Context Access & 89.7\% \\
10 & Human A-8 & 89.6\% \\
\end{longtable}

One concern is that no human participant annotated the whole dataset,
while all the LLM systems did. To see if Pilot Nr. 5 truly outperformed
all LLMs, LLM accuracies were recalculated on the subset of items that
Pilot Nr. 5 annotated. Interestingly, the LLMs' aggregate performance is
worse on this subset than on the full dataset, as shown in
Table~\ref{tbl-pilot5-vs-llms}. Pilot Nr. 5's subset may have been
particularly challenging, emphasizing this participant's expertise.
Hypothesis 1 is thus rejected based on this result.

\begin{longtable}[]{@{}
  >{\raggedleft\arraybackslash}p{(\linewidth - 4\tabcolsep) * \real{0.1000}}
  >{\raggedright\arraybackslash}p{(\linewidth - 4\tabcolsep) * \real{0.3500}}
  >{\raggedright\arraybackslash}p{(\linewidth - 4\tabcolsep) * \real{0.2000}}@{}}

\caption{\label{tbl-pilot5-vs-llms}Best human participant compared to
best LLMs when scored only on the human participant's annotated subset.}

\tabularnewline

\toprule\noalign{}
\begin{minipage}[b]{\linewidth}\raggedleft
Rank
\end{minipage} & \begin{minipage}[b]{\linewidth}\raggedright
Participant
\end{minipage} & \begin{minipage}[b]{\linewidth}\raggedright
Accuracy (subset)
\end{minipage} \\
\midrule\noalign{}
\endhead
\bottomrule\noalign{}
\endlastfoot
1 & Human Pilot Nr. 5 & \textbf{95.7\%} \\
2 & Claude Opus 4.1 & 89.9\% \\
3 & GPT-5 with Context Access & 89.1\% \\
4 & GPT-5 Context on Demand & 87.0\% \\

\end{longtable}

The accuracies of the three best-performing open-source LLMs are shown
in Table~\ref{tbl-top3-opensource-accuracy}, in which \emph{Kimi K3 with
Web Search} leads with an overall rank of 2. Researchers may prefer an
open-source model for their work if replicability is a concern, since
commercial models may covertly change their parameters and hidden
steering settings over time. Considering this very competitive
performance of \emph{Kimi K3}, open-weight models do not lag behind the
commercial ones, at least not in this task.

\begin{longtable}[]{@{}
  >{\raggedleft\arraybackslash}p{(\linewidth - 4\tabcolsep) * \real{0.1800}}
  >{\raggedright\arraybackslash}p{(\linewidth - 4\tabcolsep) * \real{0.3000}}
  >{\raggedright\arraybackslash}p{(\linewidth - 4\tabcolsep) * \real{0.2000}}@{}}

\caption{\label{tbl-top3-opensource-accuracy}Top three open-source LLMs
by accuracy.}

\tabularnewline

\toprule\noalign{}
\begin{minipage}[b]{\linewidth}\raggedleft
Overall rank
\end{minipage} & \begin{minipage}[b]{\linewidth}\raggedright
Open-source LLM
\end{minipage} & \begin{minipage}[b]{\linewidth}\raggedright
Accuracy
\end{minipage} \\
\midrule\noalign{}
\endhead
\bottomrule\noalign{}
\endlastfoot
2 & Kimi K3 with Web Search & 94.5\% \\
6 & MiMo V2.6 Pro with Web Search & 91.0\% \\
17 & GLM 5.1 & 88.3\% \\

\end{longtable}

\subsection{Hypothesis 2}\label{sec-hypothesis-2}

Hypothesis 2 posits that the highest-performing LLM outperforms the mean
human accuracy. As the human annotators differ in linguistics expertise
and English proficiency, and crowdsourcing carries a risk of inattentive
annotation, ``mean human accuracy'' can be defined in several ways.
These factors are examined first, before Hypothesis 2 is answered.

\subsubsection{Non-experts vs.~English language
experts}\label{non-experts-vs.-english-language-experts}

\begin{longtable}[]{@{}
  >{\raggedright\arraybackslash}p{(\linewidth - 2\tabcolsep) * \real{0.6000}}
  >{\raggedright\arraybackslash}p{(\linewidth - 2\tabcolsep) * \real{0.2000}}@{}}

\caption{\label{tbl-l1-vs-l2-vs-nonling}Overall accuracy of the three
human groups.}

\tabularnewline

\toprule\noalign{}
\begin{minipage}[b]{\linewidth}\raggedright
Group
\end{minipage} & \begin{minipage}[b]{\linewidth}\raggedright
Accuracy
\end{minipage} \\
\midrule\noalign{}
\endhead
\bottomrule\noalign{}
\endlastfoot
L1 Crowdworkers, English-language experts & 68.5\% \\
L2 Local Student Pilots, English-language experts & \textbf{80.8\%} \\
L1 Crowdworkers, non-experts & 69.0\% \\

\end{longtable}

Table~\ref{tbl-l1-vs-l2-vs-nonling} shows that the crowdworkers
annotated significantly less accurately than the local annotators and,
strikingly, that one group's self-identification as having expertise in
English language made almost no difference in performance as opposed to
the other. The implication is probably less that knowledge in
linguistics is not helpful for this task, but rather that Prolific's
explanation of expertise on ``English language'' did not clearly
communicate the meaning of linguistics as a discipline. For the intended
purpose of comparing student assistants to LLMs, the crowdworkers are
thus unlikely to be a suitable control group.

\subsubsection{L1 vs.~L2 speakers}\label{l1-vs.-l2-speakers}

Table~\ref{tbl-l1-vs-l2-vs-nonling} also suggests that the local
students, who were all L2 English speakers, outperform the crowdworkers.
However, a Mann-Whitney U test indicates that this difference is not
statistically significant (\emph{p} = 0.104, \emph{α} = 0.05,
two-sided). As the associated effect size was small (rank-biserial
\emph{r} = -0.39, 95\% CI {[}-0.70, -0.03{]}), a larger sample of local
students would be needed to produce a more robust result. The possible
implications are twofold: English proficiency is either not as important
as often assumed for this kind of task, and/or the sampling of local
students by their linguistics expertise was more effective than
Prolific's system of self-reported expertise\footnote{In the interest of
  transparency, it must be pointed out that two of the study pilots
  completed a course on gender and sexuality in English, giving them
  some prior exposure to the subject matter.}.

\subsubsection{Filtered human
performance}\label{sec-filtered-human-performance}

To make use of the ample data from Prolific despite the quality issues
caused by inattentiveness, they are manipulated post-hoc to better
simulate the performance of student assistants. This is done in two
ways: by taking a human majority vote and by applying performance-based
selection using context-dependent accuracy.

The human majority vote is generated by taking the most common
annotation for each data point across the Prolific-sourced,
self-identified experts in English and the local students. This
simulates a scenario in which multiple student assistants collaborate
and reach a consensus annotation for each data point. The human data
were also subjected to performance-based selection: participants with
below-chance accuracy for context-dependent items were excluded, and the
top half of the remaining participants by accuracy was retained. This
simulates a scenario in which a researcher would select the most
attentive and high-performing student assistants.

This performance-based selection yielded a group of 20 participants, of
which 2 are local students and 18 are crowdworkers. Their performance is
shown in Table~\ref{tbl-quality-based-selected-participants} below.

\begin{longtable}[]{@{}
  >{\raggedright\arraybackslash}p{(\linewidth - 4\tabcolsep) * \real{0.5000}}
  >{\raggedleft\arraybackslash}p{(\linewidth - 4\tabcolsep) * \real{0.1800}}
  >{\raggedright\arraybackslash}p{(\linewidth - 4\tabcolsep) * \real{0.1800}}@{}}
\caption{The overall accuracy of the two human groups that result from
quality-based participant
selection.}\label{tbl-quality-based-selected-participants}\tabularnewline
\toprule\noalign{}
\begin{minipage}[b]{\linewidth}\raggedright
Benchmark
\end{minipage} & \begin{minipage}[b]{\linewidth}\raggedleft
Participants
\end{minipage} & \begin{minipage}[b]{\linewidth}\raggedright
Accuracy
\end{minipage} \\
\midrule\noalign{}
\endfirsthead
\toprule\noalign{}
\begin{minipage}[b]{\linewidth}\raggedright
Benchmark
\end{minipage} & \begin{minipage}[b]{\linewidth}\raggedleft
Participants
\end{minipage} & \begin{minipage}[b]{\linewidth}\raggedright
Accuracy
\end{minipage} \\
\midrule\noalign{}
\endhead
\bottomrule\noalign{}
\endlastfoot
Human Majority Vote (all linguist humans) & 48 & \textbf{86.2\%} \\
Performance-Selected Human Mean & 20 & 78.5\% \\
\end{longtable}

With an accuracy of 94.5\%, the highest-performing LLM, \emph{Kimi K3},
outperforms all human benchmarks including the human majority vote and
the performance-selected participants in
Table~\ref{tbl-quality-based-selected-participants}. Items exclusive to
LLM-testing were disregarded for this direct comparison. Hypothesis 2 is
thus supported: the highest-performing LLM does outperform the mean
human accuracy, even when considering a more stringent benchmark of
performance-selected participants.

\subsection{Hypothesis 3}\label{sec-hypothesis-3}

Hypothesis 3 asserted that the most accurate LLM does not perform
significantly worse than the mean human on rare and innovative uses of
\emph{they} (i.e.~specific singular \emph{they}). Plural \emph{they},
the most common type, serves for comparison, and confusion matrices show
both groups' performance per category.

\begin{figure}[H]

\centering{

\pandocbounded{\includegraphics[keepaspectratio]{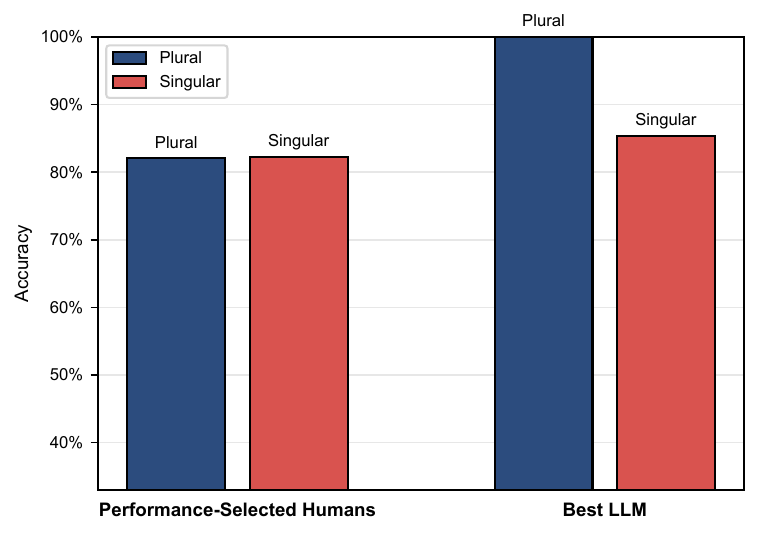}}

}

\caption{\label{fig-singular-plural-comparison}Accuracy difference
between plural/singular \emph{they} and LLMs/humans}

\end{figure}%

\begin{figure}[H]

\centering{

\pandocbounded{\includegraphics[keepaspectratio]{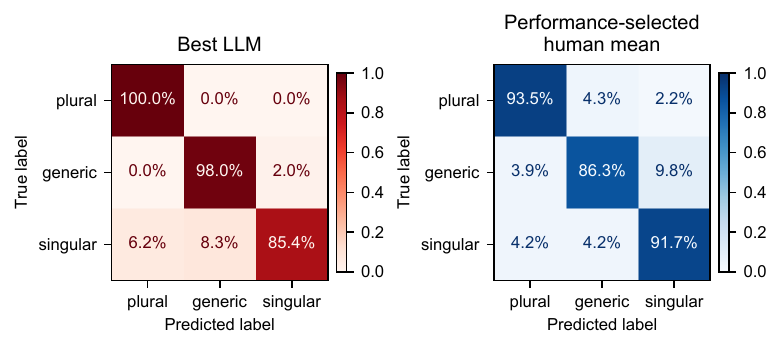}}

}

\caption{\label{fig-confusion-matrices}Confusion matrices for the best
LLM and performance-selected human mean}

\end{figure}%

Fig.~\ref{fig-singular-plural-comparison} indicates that the
singular/plural distinction did not affect humans, who had little to no
difficulty disambiguating innovative forms of \emph{they}. The best LLM,
however, performed significantly worse on singular than on plural
\emph{they} (\emph{p} = 0.038, \emph{α} = 0.05; one-sided Mann-Whitney U
test; effect size: rank-biserial \emph{r} = -1.00, 95\% CI {[}-1.00,
-1.00{]}), consistent with various observations of LLMs' bias towards
common and standard language (Pan et al., 2025; Zhou et al., 2025).
Still, its performance on singular \emph{they} was not significantly
worse than the performance-selected human mean, as shown by two
one-sided \emph{t}-tests (TOST) for non-inferiority (\emph{p\_lower} =
0.037, \emph{α} = 0.05, Δ = 0.02), supporting Hypothesis 3.

The confusion matrices in Fig.~\ref{fig-confusion-matrices} again show
singular \emph{they} as the best LLM's most challenging category,
whereas humans erred most with generic \emph{they}. The participants'
comments suggest that one source of confusion is the referents
\emph{everyone} and \emph{no one}, which the instructions explicitly
label as plural, along with \emph{nobody}. Many participants instead
seem to have retained the rule that statements ``that {[}apply{]} to all
X'' are generic, or based their judgment on the singular verb agreement,
leading me to revise the typology for future work.\footnote{In a
  post-hoc analysis, counting \emph{generic} responses as correct on all
  three items that involve such antecedents raised the
  performance-selected human mean from 78.5\% to 79.7\% (+1.1 percentage
  points).} This reveals a key difference between human and LLM
annotators: LLMs tend to excel at reproducing instructed patterns even
when the instructions are internally inconsistent, whereas humans often
infer the underlying logic, potentially drawing on their past education,
and may override inconsistent instructions. In a real-world annotation
setting, participants might have asked for clarification, thereby
exposing inconsistencies and improving the annotation scheme. Their
quantitative weakness can thus also be interpreted as a qualitative
strength.

\subsection{Additional metrics}\label{sec-additional-metrics}

Three further analyses complete the comparison of the two groups:
performance on context-(in)dependent items, a projection onto
naturalistic data, and cost.

\subsubsection{Context (in-)dependence}\label{sec-context-dependence}

\begin{figure}[H]

\centering{

\pandocbounded{\includegraphics[keepaspectratio]{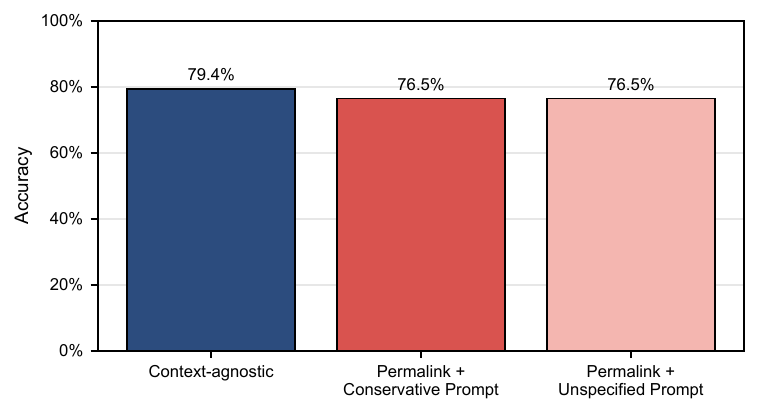}}

}

\caption{\label{fig-gpt5-run-accuracies}Three different runs of the same
\emph{GPT-5} model, exclusively on context-dependent data, with
different context access and prompt configurations. Dark bars indicate
models that were unable to access the online context, lighter bars
indicate models that were able to. Of the lighter bars, the conservative
prompt told the model to be cautious and to access context less
liberally (medium shade) compared to the unspecified prompt (light
shade)}

\end{figure}%

Fig.~\ref{fig-gpt5-run-accuracies} displays the run-level accuracies for
the three \emph{GPT-5} configurations: context-agnostic, permalink +
conservative prompt, and permalink + unspecified prompt. It is notable
that the two setups with access to the items' discourse-context from
Reddit performed worse than the context-agnostic one, reinforcing the
findings of Shore et al. (2025) that models struggle with the complexity
of such a combined task.

Closer analysis of results from the conservative prompting of
\emph{GPT-5} with full URL access shows that of the 34 items marked as
context-dependent, less than half (14) successfully triggered the model
to access the context. In turn, of the 25 items that did cause the model
to check context, I had only marked 14 as context-dependent. Among the
items that the model had considered context-independent and I had
considered context-dependent, however, the model's accuracy still lies
at a relatively high 80.0\% -- very close to the average accuracy of the
local study pilots over all items at 80.8\%. Among the items for which
the model had correctly decided to access context, its accuracy was only
76.0\%, indicating that it struggles with heterogeneous and complex
contexts. This is precisely the boundary that Morin \& Marttinen Larsson
(2026, p. 31) anticipate for their own \emph{GPT-5} pipeline, which
classified concordance lines of 15 words on either side of the target
rather than entire discussion threads: tasks ``requiring extremely
nuanced semantic distinctions, extensive discourse context, or
specialised domain knowledge may prove challenging even with careful
prompt engineering.''

When the model is not instructed to be conservative with context access,
it accordingly demonstrates a clear tendency to access it more often: 28
of the 34 context-dependent cases were checked for context, indicating a
high sensitivity (i.e.~recall) of 82\%. However, the model checked a
total of 106 cases for context, of which only 28 were actually
context-dependent, yielding a precision of only 26\%. This means that
while the model is relatively good at correctly identifying cases that
require context, it also checks for context in many cases in which the
context can actually introduce unnecessary information that leads to
confusion. In this study, the accuracy decreased by 10\% (from 94\% to
84\%) when the model accessed context vs.~when it did not, hence the
decrease in accuracy observed in Fig.~\ref{fig-gpt5-run-accuracies} for
the two runs that had access to context.

This analysis reveals that the model's behavior does not fully align
with my expectations regarding which items require context for
disambiguation. It is not clear exactly what causes this difference in
behavior, but it does suggest that LLMs' annotation capabilities differ
from those of humans, generally to the detriment of the results. In
terms of study design, it also suggests that LLMs fare better without
researching information on their own in order to complete a task, as
correctly identifying ambiguity seems to remain too great a challenge
for such models. Their applicability may thus be limited in the study of
especially complex linguistic phenomena that require multi-step
decision-making.

\subsubsection{Projection onto naturalistic
data}\label{sec-naturalistic-projection}

\begin{longtable}[]{@{}
  >{\raggedright\arraybackslash}p{(\linewidth - 4\tabcolsep) * \real{0.4500}}
  >{\raggedright\arraybackslash}p{(\linewidth - 4\tabcolsep) * \real{0.2500}}
  >{\raggedleft\arraybackslash}p{(\linewidth - 4\tabcolsep) * \real{0.2500}}@{}}

\caption{\label{tbl-naturalistic-projection}Naturalistic projection
using class frequencies from old annotated data (file
\texttt{ma-era-wth-lores.xlsx} in the study repository).}

\tabularnewline

\toprule\noalign{}
\begin{minipage}[b]{\linewidth}\raggedright
Group
\end{minipage} & \begin{minipage}[b]{\linewidth}\raggedright
Projected accuracy
\end{minipage} & \begin{minipage}[b]{\linewidth}\raggedleft
Projected MCC
\end{minipage} \\
\midrule\noalign{}
\endhead
\bottomrule\noalign{}
\endlastfoot
Performance-selected human mean & 92.3\% & 0.809 \\
Best LLM & 98.6\% & 0.961 \\

\end{longtable}

Using naturalistic class frequencies from the original source data of
previous projects (Hartmann, 2022 and unpublished material) (\emph{N} =
1,116), projected accuracy remains high for both approaches: 92.3\% for
the performance-selected human mean and 98.6\% for the best LLM. The
corresponding projected MCC values are 0.809 (humans) and 0.961 (LLM),
indicating very strong overall quality for both when class balance is
made more realistic. The non-negligible superiority of the LLMs' MCC
value derives from their perfect performance on plural \emph{they}
items, which are the most frequent class in naturalistic data -- though
the MCC is designed to correct for class imbalance, this difference is
still apparent. This suggests that LLMs have a clear advantage in common
items and thus reach better overall scores. However, research must also
satisfactorily address edge cases; these results should not be
over-interpreted as a blanket endorsement of LLMs over humans, but
rather as a hint towards how humans and LLMs can collaborate: in a
complementary fashion, the latter can process the multitudes of
easily-classified and common instances and the former can be left with
the more challenging and rare cases, which are often the most relevant
ones in linguistics research. This approach leverages the strengths of
both humans and LLMs while mitigating their respective weaknesses.

\subsubsection{Cost}\label{sec-cost}

The cost of crowdworkers is difficult to estimate because of their
inattentiveness and the overall quality of their annotations: 23 out of
41 crowdworkers, or 56.1\%, had to be disregarded. Nominally, the
crowdworkers cost \$40.98 per 100 examples. However, since all of them
had to be paid but only the responses of the retained crowdworkers were
used, the effective cost rises to \$93.33 per 100 usable examples. This
illustrates a risk of crowdsourcing: inadequate vetting can cause
significant inefficiencies and costs. By comparison, the local student
pilots cost \$68.75 per 100 examples, more than the crowdworkers'
nominal cost. In exchange, the locals annotated more accurately than the
crowdworkers who self-identified as English-language experts (80.8\%
vs.~68.5\%; see Table~\ref{tbl-l1-vs-l2-vs-nonling}), although this
difference was not statistically significant given the small pilot
sample. Measured per 100 correctly annotated examples, the crowdworkers
were cheaper (\$59.82 vs.~\$85.13), but per usable example, the
carefully recruited local students were the cheaper option.

Humans needed one to two hours per 100 examples and can work only about
eight hours a day, whereas LLM inference, though untimed here, takes
seconds to minutes per item, around the clock. The ``cost'' of LLMs is
contentious: though far less was spent on them than on humans, concrete
values would be misleading. Firstly, AI companies are currently
providing their services at a considerable loss (see Ren et al., 2024,
pp. 4--5), so usage through a university-hosted cluster would cost more
than was spent here. Secondly, cost does not equal pricing; the former
expresses the more fundamental consequences of resource allocation and
of deploying such models. Methodological work nonetheless tends to treat
API pricing as the cost of annotation: Morin \& Marttinen Larsson (2026,
p. 4) present the \$104 they spent in API fees as showing that
large-scale annotation ``is financially viable even for modestly funded
studies'', and weigh this against research-assistant wages (2026, p.
29). There are also energy costs and the broader environmental and
ethical considerations of large-scale computing. Morin \& Marttinen
Larsson (2026, p. 30) counter, citing Ren et al. (2024), that LLM
inference uses far less energy, water, and carbon than equivalent human
labor. Ren et al.'s estimates, however, were made for a mid-sized,
non-reasoning model writing a 500-word text, and they attribute an
amortized share of a human's entire footprint to the task, an approach
the authors themselves concede (2024, p. 4) ``may seem unconventional''.
They also caution (2024, p. 4) that with growing model sizes, LLMs ``may
be more energy-consuming than human labor''. Future projects employing
LLMs should therefore transparently report how they address the issue of
true cost (see Section~\ref{sec-conclusions}).

\section{Discussion}\label{sec-discussion}

Hypothesis 1, that an LLM would outperform the highest-ranked human, was
rejected due to one especially strong human participant (who
outperformed even the study author and their peer considered in IAA
testing; Section~\ref{sec-hypothesis-1}). Hypotheses 2 and 3, that LLMs
outperform any generalized human metric (Hypothesis 2;
Section~\ref{sec-hypothesis-2}) despite the difficulty with innovative
language (Hypothesis 3; Section~\ref{sec-hypothesis-3}), were supported
by \emph{Kimi K3}'s superior accuracy even when compared to the
highest-performing humans. The cost and speed comparison
(Section~\ref{sec-cost}) may seem like the final nail in the coffin for
the human assistant, as LLMs appear far cheaper and faster. But not only
should we be careful about not confusing pricing with actual cost, the
results also indicated the strength of complementary human--LLM
approaches. Namely, the human participants' apparent weakness was
revealed to be a valuable strength: their ``underperformance'' was
related to an inconsistency in the annotation scheme. The human
reasoning that revealed this inconsistency ultimately improved the
methodology, a common outcome in cases of initial low agreement scores
(see Aroyo \& Welty, 2015, among other work by Aroyo and associates).
LLMs, on the other hand, were simply very good at executing instructions
exactly as formulated, inconsistencies notwithstanding. Morin \&
Marttinen Larsson (2026, p. 31) likewise insist that their pipeline
``requires human researchers to conceptualise the annotation task,
recognise relevant edge cases, interpret validation results, and take
responsibility for data quality.'' The present results suggest that this
conceptual work does not rest with the researcher alone: the annotators'
reasoning contributed to it as well.

\section{Limitations}\label{limitations}

Study results are interpreted in light of the following limitations,
which are grouped into three categories: internal validity, external
validity, and operational validity.

\subsection{Internal validity}\label{internal-validity}

The first limitation to internal validity is that the inferential
analyses were primarily based on aggregated accuracies rather than a
full item-level hierarchical model. Residual item-difficulty effects
(i.e.~some items may be inherently easier or harder than others) may
thus have influenced results. However, this reflects data retrieval work
in linguistics where examples from natural language show strong
variation. Fully ambiguous cases were not included in this study, which
continue to pose a challenge to LLMs when simultaneously instructed to
detect ambiguity and classify unambiguous items. The removal of 31\% of
the items in the dataset for length may also have simplified the task
for LLMs, which are known to be sensitive to input length. However, the
pronoun-antecedent distance across investigated written genres is on
average shorter than three sentences away from their antecedent (Aktaş
\& Stede, 2020, Section 4.4; Aktaş \& Stede, 2022, Section 2), so this
limitation may not be very consequential.

\subsection{External validity}\label{external-validity}

As with all studies, these results are limited in their generalizability
to other contexts. Natural languages are in a state of constant
evolution; some aspects like gender-inclusive language change
particularly quickly, as does LLM-based technology. Researchers applying
LLM-based annotation protocols must consider the unique qualities of
each model, annotation scheme and task, and dataset and adapt their
methods accordingly. Furthermore, context-sensitivity was only tested
for LLMs and not for humans for the sake of experiment brevity. It could
be worthwhile to study human annotators' annotation performance when
faced with ambiguous data.

\subsection{Operational validity}\label{operational-validity}

A valuable test of operational validity would be to design a human--LLM
cooperative approach and compare its performance to that of each group
alone, especially given the complementary strengths and weaknesses of
humans and LLMs in this study. Public discourse often asserts that human
labor is not threatened by AI as long as they work together (i.e.~the
human-in-the-loop, or HITL, principle), a claim worth testing in a
scientific setting. It is also important to acknowledge that confidence
scores are not available when operationalizing LLMs for data retrieval
(unlike in previous approaches to most NLP tasks), meaning that
researchers cannot manipulate recall/precision trade-offs when
processing large corpora. Given that multi-step reasoning remains a
challenge for LLMs, prompt design must also be rigorously tested and
piloted before being applied to large corpora.

\section{Conclusions}\label{sec-conclusions}

Results indicate that LLMs have largely reached human-level performance
in complex annotation tasks (see Section~\ref{sec-discussion}). For
corpus retrieval tasks like this one, four lessons emerge: open-weight
and transparent models facilitate reproducibility; processing steps
should be separated (ambiguity assessment first, then classification);
performance should be validated on data samples at each stage (see Morin
\& Marttinen Larsson, 2026, pp. 10--14); and human oversight remains
crucial for quality and for edge cases (see Yu et al., 2024, Section 5),
often the most crucial part of the research. Open Science further lets
the research community see and address potential shortcomings.

The growing role of LLMs in research gives these findings clear ethical
implications for linguistics and the humanities more generally. LLMs'
pricing and speed advantages affect research funding, which dictates our
field as much as any other and may not sufficiently incentivize
environmentally friendly research design. Morin \& Marttinen Larsson
(2026, p. 29) even call on funding bodies to ``recognise API access as
essential research infrastructure''. LLM use should indeed be budgeted
for, but university-hosted infrastructure running open-weight models
would be the more sustainable investment, also addressing Morin and
Marttinen Larsson's own data-sovereignty concerns (2026, p. 30). The
``cost'' of LLMs is far-reaching, not least environmentally, while the
value of humans is often unforeseeable: they improve research in small
but crucial ways, as they did in this study in several ways, including
the methodology's underlying logic and not least through excellent
copy-editing.

My central claim is that the humanities can indeed profit from AI, but
only through ethical and responsible use. True intelligence, a deeply
complex and nuanced concept, encompasses more than pattern prediction
(Pope Leo XIV, 2026, Sections 99, 100). The social sciences and
humanities are especially well-suited to measure the impact of new
technologies and shape them in a socially responsible way (2026, Section
23). Morin \& Marttinen Larsson (2026, p. 4) argue that their approach
``redefines the linguist's role in the design and execution of research
projects'', counting junior scholars among its main beneficiaries (2026,
p. 30): true, perhaps, for junior scholars leading their own projects,
but not for the student assistants whose work is being automated.
Replacing human labor with AI may appear cost-efficient but comes at the
hidden cost of depriving early-career knowledge workers of foundational
skill development and practical training. If LLMs are to become
``copilots for linguists'' (Torrent et al., 2023), the next generation
of pilots still needs to be trained. Redefining the role of student
assistants, for instance to suit the new digital reality and leverage
the strengths of both humans and AI, will ultimately improve research
outcomes and the quality of scholarly work. AI is only ever as powerful
as the human intelligence it extends. At least for now.

\section*{Statements and
Declarations}\label{statements-and-declarations}
\addcontentsline{toc}{section}{Statements and Declarations}

\textbf{Funding} This study received no third-party funding. It was
funded by the English Department of the University of Zurich as part of
the author's employment.

\textbf{Competing interests} The author has no relevant financial or
non-financial interests to disclose.

\textbf{Ethics approval} The study did not undergo review by an ethics
committee. It involved no health-related data or interventions and thus
falls outside the scope of the Swiss Human Research Act (HRA, SR
810.30), which governs mandatory ethics review in Switzerland. On the
handling of the Reddit data, see Section~\ref{sec-dataset}.

\textbf{Consent to participate} All participants were informed about the
content of the task and its potential risks before taking part: the
local participants by e-mail, the crowdworkers through the Prolific
study description. All items had been screened by the author beforehand;
the only identified risk was exposure to controversial political
opinions expressed in potentially uncomfortable ways. Participation was
voluntary, and participants could withdraw at any time.

\textbf{Data availability} The dataset (the Reddit items with their
ground-truth labels), all human and LLM annotations, the prompts, and
the survey materials are available in the study repository on OSF at
https://doi.org/10.17605/OSF.IO/VEB2J (Hartmann, 2026b). The links to
the original comments, which repeating the runs with context access
requires, are available from the author for research purposes on
request.

\textbf{Code availability} The analysis code and the source of this
article are available in the same repository.

\newpage

\section*{References}\label{references}
\addcontentsline{toc}{section}{References}

\phantomsection\label{refs}

\end{document}